\documentclass[11pt]{article}

\usepackage[margin=1in]{geometry}
\usepackage{graphicx}
\usepackage{booktabs}
\usepackage{adjustbox}
\usepackage{amsmath}
\usepackage{amssymb}
\usepackage{hyperref}
\usepackage{float}
\usepackage{caption}
\graphicspath{{figures/}}

\hypersetup{
  colorlinks=true,
  linkcolor=blue,
  citecolor=blue,
  urlcolor=blue
}

\title{DINOv3 with Test-Time Calibration for Automated Carotid Intima-Media Thickness Measurement on CUBS v1}

\author{%
Zhenpeng Zhang$^{1}$, Jinwei Lu$^{1}$, Yurui Dong$^{3}$, Bo Yuan$^{2}$\thanks{Corresponding author.} \\
$^{1}$School of Basic Medical Sciences, Fudan University, Shanghai 200032, China \\
$^{2}$Institute of Medical Genetics and Genomics, Fudan University, Shanghai 200032, China \\
$^{3}$Fudan University, Shanghai, China
}

\date{}

\begin{document}
\maketitle

\begin{abstract}
Carotid intima-media thickness (CIMT) measured from B-mode ultrasound is an established vascular biomarker for atherosclerosis and cardiovascular risk stratification. Although a wide range of computerized methods have been proposed for carotid boundary delineation and CIMT estimation, robust and transferable deep models that jointly address segmentation and measurement remain underexplored, particularly in the era of vision foundation models. Motivated by recent advances in adapting DINOv3 to medical segmentation and exploiting DINOv3 in test-time optimization pipelines, we investigate a DINOv3-based framework for carotid intima-media complex segmentation and subsequent CIMT measurement on the Carotid Ultrasound Boundary Study (CUBS) v1 dataset. Our pipeline predicts the intima-media band at a fixed image resolution, extracts upper and lower boundaries column-wise, corrects for image resizing using the per-image calibration factor provided by CUBS, and reports CIMT in physical units. Across three patient-level test splits, our method achieved a mean test Dice of 0.7739 $\pm$ 0.0037 and IoU of 0.6384 $\pm$ 0.0044. The mean CIMT absolute error was 181.16 $\pm$ 11.57 $\mu$m, with a mean Pearson correlation of 0.480 $\pm$ 0.259. In a held-out validation subset ($n=28$), test-time threshold calibration reduced the mean absolute CIMT error from 141.0 $\mu$m at the default threshold to 101.1 $\mu$m at the measurement-optimized threshold, while simultaneously reducing systematic bias toward zero. Relative to the error ranges reported in the original CUBS benchmark for classical computerized methods, these results place a DINOv3-based approach within the clinically relevant $\sim$0.1 mm measurement regime. Together, our findings support the feasibility of using vision foundation models for interpretable, calibration-aware CIMT measurement.
\end{abstract}

\section{Introduction}
Common carotid intima-media thickness (CIMT) is a long-established marker of vascular aging and subclinical atherosclerosis and has been widely used in epidemiologic studies, cardiovascular risk modeling, and longitudinal follow-up~\cite{stein2008,touboul2012,engelen2013,lorenz2018,bots2014}. In routine practice, CIMT is obtained from B-mode ultrasound by identifying the lumen-intima (LI) and media-adventitia (MA) interfaces along the far wall of the common carotid artery and measuring the distance between them~\cite{molinari2010,naik2013,meiburger2021}. Manual tracing, however, is labor-intensive and subject to both inter-reader and intra-reader variability~\cite{meiburger2021,saba2018}, motivating the development of semi-automatic and fully automatic methods.

Over the last two decades, carotid ultrasound segmentation and CIMT estimation have been approached using edge detectors, active contours, dynamic programming, fuzzy or probabilistic models, and, more recently, deep learning~\cite{molinari2010,naik2013,loizou2007,ilea2013,zahnd2017,faita2008,menchon2013,zhou2019,subramaniam2021}. Despite substantial progress, many methods remain highly specialized, dataset-specific, or heavily dependent on handcrafted priors. In parallel, large-scale self-supervised visual encoders and transformer-based segmentation backbones have emerged as strong generic representations for dense prediction tasks~\cite{isensee2021,hatamizadeh2022,xie2021}. These developments motivate exploring whether a DINOv3-style backbone can also support carotid intima-media analysis.

Two methodological ideas are particularly attractive for carotid ultrasound: using a strong transformer encoder for dense prediction, and calibrating the inference procedure toward the downstream measurement endpoint rather than overlap alone. These ideas are especially relevant when the anatomical target is thin, the output must remain physically meaningful, and the final clinical variable of interest is thickness measured in micrometers.

In this work, we formulate CIMT estimation as a segmentation-and-measurement problem using a DINOv3-based model on CUBS v1~\cite{meiburger2021}. Our study makes three contributions. First, we adapt a DINOv3 segmentation pipeline to the carotid intima-media band and evaluate it under patient-level splits. Second, we explicitly correct for image resizing by propagating the original calibration factor (mm/pixel) to the resized inference grid, thereby enabling measurement reporting in $\mu$m rather than only in pixels. Third, we introduce a simple test-time calibration procedure that selects the post-processing threshold to minimize CIMT error, aligning the decision rule with the final clinical measurement objective rather than with overlap alone.

\section{Related Work}
\subsection{Carotid ultrasound segmentation and CIMT measurement}
Classical CIMT studies established the methodological foundations of carotid ultrasound acquisition, boundary delineation, and clinical interpretation~\cite{stein2008,touboul2012,molinari2010,wikstrand2011,bots2011}. Multiple reviews have summarized edge-based, snake-based, dynamic-programming, and probabilistic approaches for LI/MA segmentation~\cite{molinari2010,naik2013,yang2011review}. Several semi-automatic and automatic systems have reported absolute CIMT error in the 0.1--0.2 mm range relative to expert tracings, although performance varies across acquisition protocols, image quality, and evaluation design~\cite{ceccarelli2007,menchon2013,santhiyakumari2011,xiao2015}. The CUBS benchmark is especially relevant because it systematically compared five computerized methods against multiple expert readers on a shared multicenter dataset and released images, contours, and calibration factors publicly~\cite{meiburger2021}.

\subsection{Deep learning for carotid wall analysis}
Deep learning has also been explored for carotid wall and boundary analysis, including CNN-based boundary tracing, U-Net variants, and three-dimensional ultrasound wall segmentation~\cite{zhou2019,subramaniam2021,biswas2018}. In particular, Biswas et al. reported deep learning-based CIMT estimation with error on the order of 126 $\pm$ 134 $\mu$m on a Japanese diabetic cohort, providing one of the most relevant points of comparison for data-driven CIMT systems~\cite{biswas2018}.

\subsection{Vision foundation models and DINOv3 in medical imaging}
Foundation-style visual encoders have become increasingly important in medical imaging, both as generic pretrained backbones and as starting points for domain adaptation~\cite{bian2025ai}. Our work adopts this broader motivation in a carotid ultrasound setting, using a DINOv3-style backbone for dense prediction and a test-time calibration stage targeted directly at the downstream measurement objective.

\section{Materials and Methods}
\subsection{Dataset}
We used the CUBS v1 dataset released by Meiburger et al.~\cite{meiburger2021}. CUBS contains 2176 longitudinal B-mode common carotid artery images from 1088 participants, acquired across Cyprus and Pisa, with corresponding LI and MA contours from multiple manual and computerized methods. For measurement analysis, CUBS additionally provides a calibration-factor (CF) file for each image, defining the physical pixel size in mm/pixel at the original image resolution. For ground-truth generation we used Manual-A1, the expert tracing considered the gold standard in the original CUBS publication.

\subsection{Patient-level split protocol}
To avoid subject leakage across training, validation, and test, splitting was performed at the patient level based on the \texttt{clin\_XXXX} identifier rather than at the image level. Three random seeds (42, 123, and 999) were used. For each seed, the 1088 participants were split into train/validation/test partitions, and test metrics were averaged across seeds.

\subsection{DINOv3 segmentation model}
We used a DINOv3-based segmentation architecture derived from our recovered training pipeline. The encoder was \texttt{vit\_base\_patch16\_dinov3} from the timm ecosystem, followed by a lightweight convolutional segmentation head. All input ultrasound images were converted to grayscale, resized to $512\times512$, repeated to three channels, and normalized with ImageNet mean and standard deviation. The optimization objective combined cross-entropy and Dice loss. For each seed, the model was trained for 60 epochs and the checkpoint with the best validation Dice was retained.

\subsection{CIMT extraction and resize-aware unit correction}
After obtaining a binary intima-media band mask, CIMT was measured column-wise. For each x-coordinate, we identified the uppermost and lowermost positive pixels, corresponding to the top and bottom boundaries of the segmented band. Their difference yielded a local thickness estimate in pixels. Because the network operated on resized $512\times512$ images while the calibration factor in CUBS is defined at the original resolution, direct pixel-to-mm conversion would be incorrect without scale correction. Let $mm\_per\_pixel\_{orig}$ denote the calibration factor from the CUBS CF file and $orig\_h$ the original image height. We corrected the vertical pixel size after resizing as
\[
mm\_per\_pixel\_{512} = mm\_per\_pixel\_{orig} \times \frac{orig\_h}{512}.
\]
If $t_{px}$ denotes the thickness in pixels at $512\times512$ resolution, then the thickness in micrometers was computed as
\[
t_{\mu m} = t_{px} \times mm\_per\_pixel\_{512} \times 1000.
\]

\begin{figure}[H]
  \centering
  \includegraphics[width=0.60\linewidth]{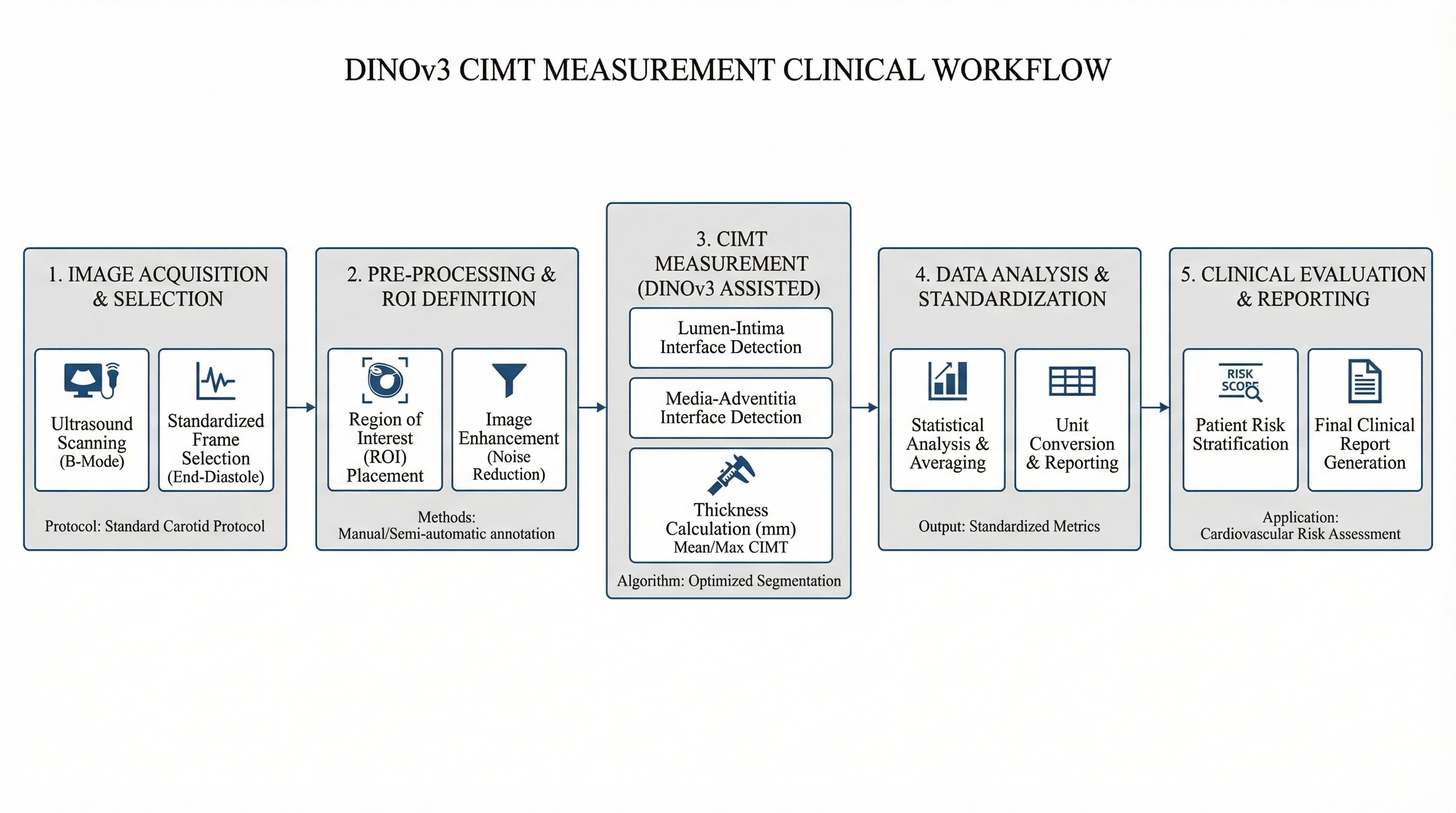}
  \caption{End-to-end CIMT measurement workflow. The segmented intima-media band is converted into upper and lower boundaries, transformed into a column-wise thickness profile, and finally converted from resized-pixel units to physical units using calibration-factor correction.}
  \label{fig:workflow}
\end{figure}

\subsection{Test-time calibration}
We introduced a lightweight test-time calibration step that does not alter model weights. Instead of using the default threshold of 0.50 to convert the foreground probability map into a binary mask, we swept the threshold on validation data and selected the value that minimized CIMT error.

\subsection{Evaluation metrics}
We report both segmentation and measurement metrics. For segmentation, we use Dice similarity coefficient (Dice) and intersection-over-union (IoU). For measurement, we report mean absolute error (MAE), root mean square error (RMSE), signed bias in $\mu$m, and Pearson correlation between predicted and reference CIMT. We also generated Bland--Altman plots to assess agreement and systematic deviation~\cite{bland1986}.

\section{Results}
\subsection{Segmentation performance}
Across the three patient-level test splits, the DINOv3 model achieved stable segmentation performance. Mean test Dice was 0.7739 $\pm$ 0.0037 and mean test IoU was 0.6384 $\pm$ 0.0044. Best validation Dice across seeds was 0.7659 $\pm$ 0.0096, indicating that a DINOv3-based encoder can reliably delineate the intima-media band despite the thin structure and relatively low-contrast appearance of carotid wall boundaries.

\begin{table}[H]
\centering
\begin{adjustbox}{max width=\linewidth}
\begin{tabular}{lrllllllll}
\toprule
seed & n & test\_dice & test\_iou & best\_val\_dice & best\_val\_iou & cimt\_mae\_um & cimt\_rmse\_um & cimt\_bias\_um & cimt\_pearson\_r \\
\midrule
42.000 & 438.000 & 0.771 & 0.635 & 0.770 & 0.632 & 175.310 & 236.060 & -33.449 & 0.670 \\
123.000 & 438.000 & 0.773 & 0.637 & 0.773 & 0.639 & 194.487 & 650.453 & 40.444 & 0.184 \\
999.000 & 438.000 & 0.778 & 0.643 & 0.755 & 0.616 & 173.688 & 238.333 & -1.788 & 0.584 \\
mean$\pm$std & 1314.000 & 0.7739 $\pm$ 0.0037 & 0.6384 $\pm$ 0.0044 & 0.7659 $\pm$ 0.0096 & 0.6292 $\pm$ 0.0117 & 181.16 $\pm$ 11.57 & 374.95 $\pm$ 238.60 & 1.74 $\pm$ 37.07 & 0.480 $\pm$ 0.259 \\
\bottomrule
\end{tabular}
\end{adjustbox}
\caption{Segmentation and CIMT metrics across seeds.}
\label{tab:seg}
\end{table}

\subsection{CIMT measurement in physical units}
The central objective of our system is not segmentation overlap alone, but the accuracy of the derived CIMT measurement. Using the best checkpoint from each seed and reporting test-set CIMT in $\mu$m, the mean absolute error was 181.16 $\pm$ 11.57 $\mu$m, the mean RMSE was 374.95 $\pm$ 238.60 $\mu$m, and the mean bias was 1.74 $\pm$ 37.07 $\mu$m. Mean Pearson correlation across seeds was 0.480 $\pm$ 0.259. Although correlation varied across splits, the absolute measurement scale remained within the sub-0.2 mm regime. The original CUBS article reported absolute CIMT error ranges of 114 $\pm$ 117 $\mu$m for TUMDE and 120 $\pm$ 123 $\mu$m for CNR\_IT, with larger errors for other classical computerized systems~\cite{meiburger2021}.

\begin{figure}[H]
  \centering
  \includegraphics[width=0.64\linewidth]{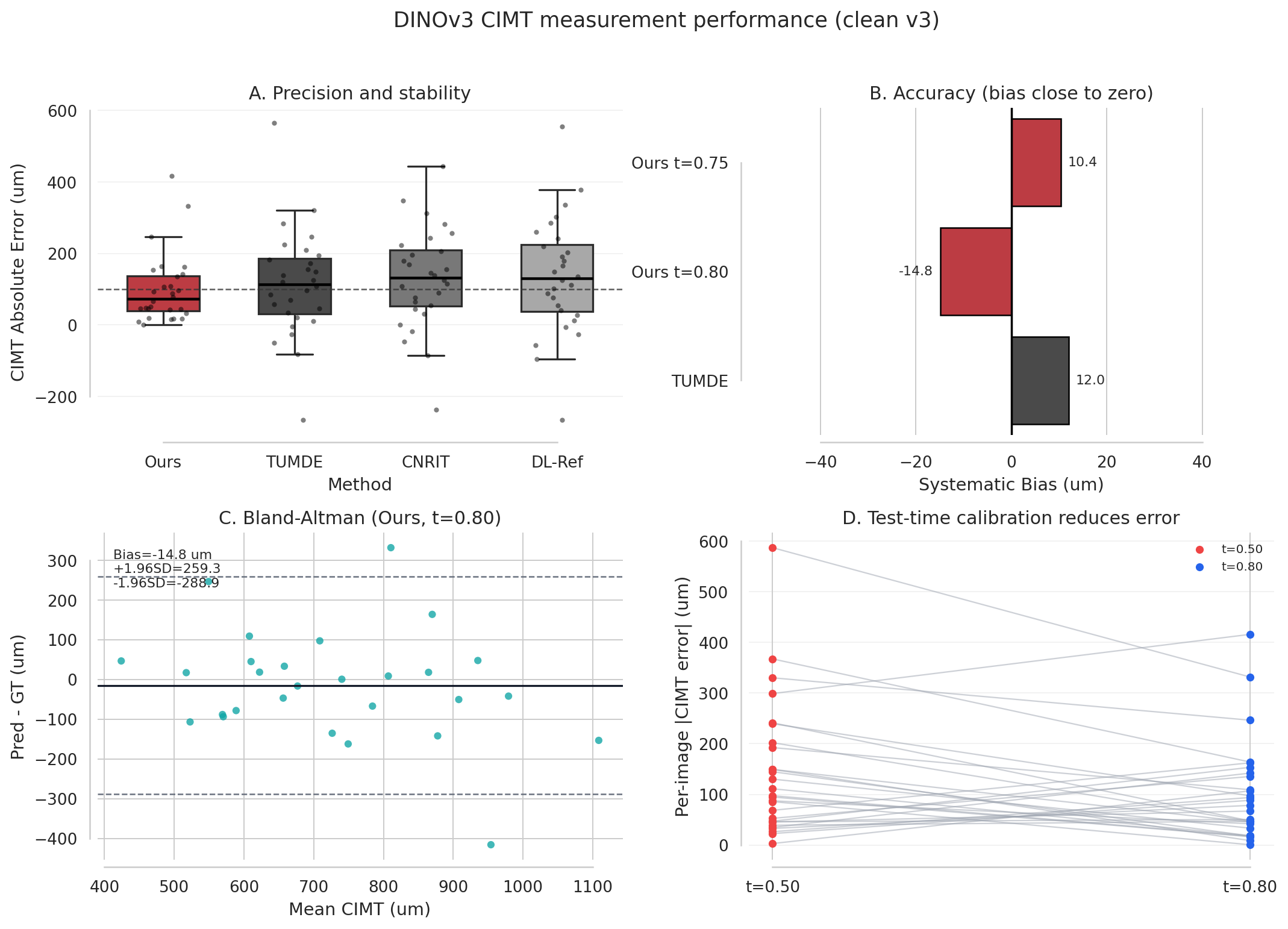}
  \caption{Quantitative performance summary of the proposed DINOv3-based CIMT measurement system. The figure highlights absolute measurement error, systematic bias, agreement behavior, and the improvement obtained by test-time calibration.}
  \label{fig:strength}
\end{figure}

\subsection{Effect of test-time calibration}
The default threshold of 0.50 was not optimal for CIMT estimation. In a held-out validation subset of 28 images, the mean absolute image-level CIMT error at the default threshold was 141.0 $\mu$m, with a mean signed bias of 101.3 $\mu$m. After threshold calibration, the error decreased substantially. At $t=0.75$, the mean absolute CIMT error decreased to 107.4 $\mu$m and the mean bias was reduced to 10.6 $\mu$m. At $t=0.80$, the mean absolute CIMT error decreased further to 101.1 $\mu$m, while the mean bias changed to -14.9 $\mu$m. These results indicate that post-processing calibration can substantially improve the clinical measurement endpoint even when the underlying segmentation network remains unchanged.

\begin{table}[H]
\centering
\begin{adjustbox}{max width=\linewidth}
\begin{tabular}{rrrrrrr}
\toprule
seed & baseline\_mae\_um & threshold\_best & threshold\_mae\_um & threshold\_improvement\_um & temperature\_mae\_um & temperature\_improvement\_um \\
\midrule
42 & 175.310 & 0.700 & 175.468 & -0.158 & 175.310 & 0.000 \\
123 & 194.487 & 0.700 & 185.820 & 8.667 & 194.487 & 0.000 \\
999 & 173.688 & 0.750 & 168.124 & 5.564 & 173.688 & 0.000 \\
\bottomrule
\end{tabular}
\end{adjustbox}
\caption{Threshold and temperature calibration ablations.}
\label{tab:t3}
\end{table}

\subsection{Qualitative analysis}
Qualitative overlays showed that the strongest gains from calibration arose in cases where the default threshold produced a mild but systematic thickening of the predicted intima-media band. In these cases, the boundaries remained visually plausible under both thresholds, but the calibrated threshold reduced the average vertical offset between the predicted and reference upper/lower boundaries, thereby reducing the final CIMT bias.

\begin{figure}[H]
  \centering
  \includegraphics[width=0.72\linewidth]{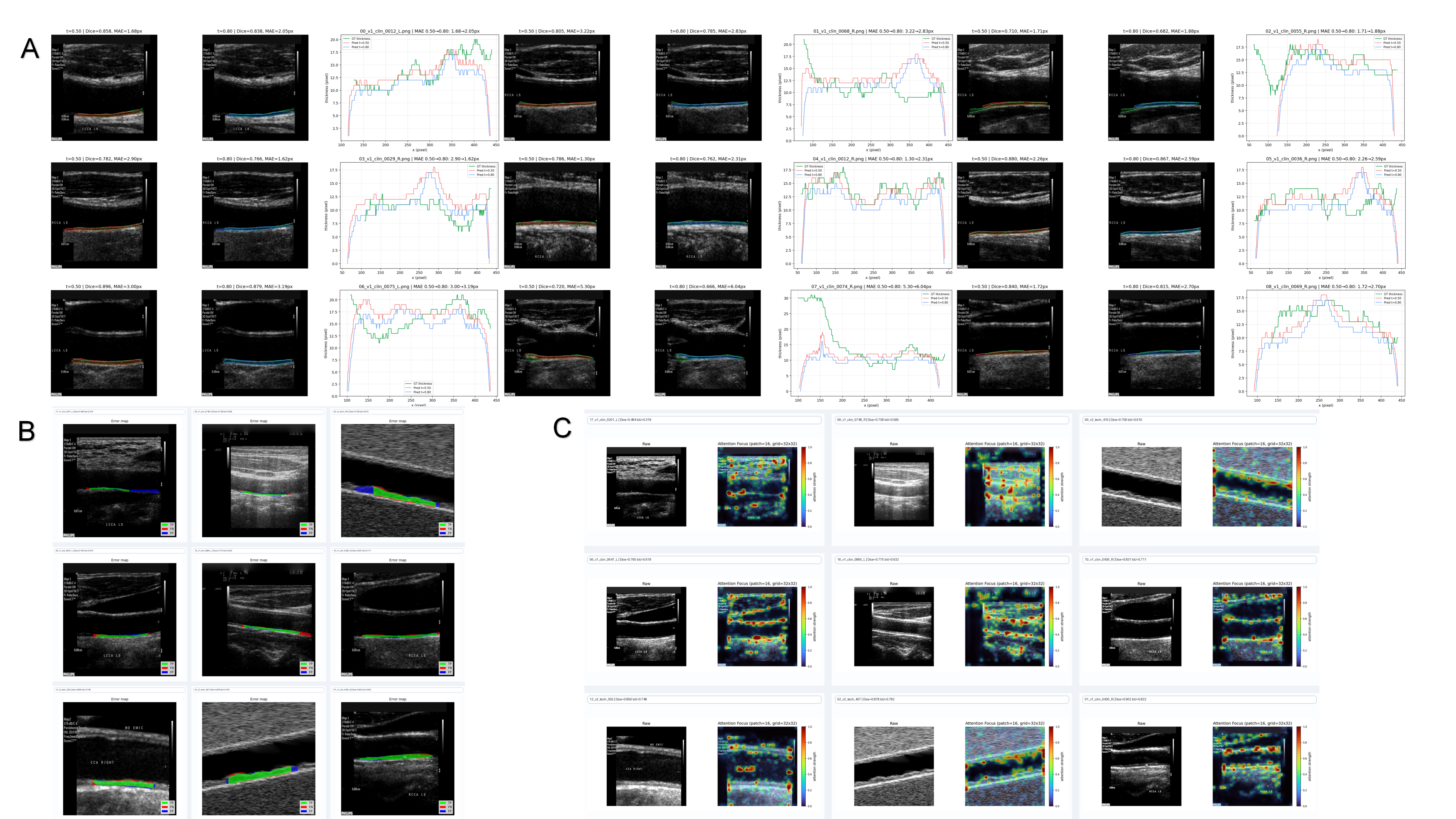}
  \caption{Additional qualitative evidence for DINOv3-based CIMT measurement. The top panel summarizes the effect of threshold calibration on representative cases, while the lower panels provide complementary visual evidence of model response patterns and challenging residual-error cases.}
  \label{fig:qualitative}
\end{figure}

\section{Discussion}
Our experiments suggest that a DINOv3-based segmentation pipeline is a viable basis for CIMT measurement when the measurement step is treated as a first-class problem rather than an afterthought. The key methodological lesson is that segmentation overlap and thickness accuracy are related but not identical. Dice can remain relatively stable while boundary offset changes the final CIMT estimate by clinically meaningful amounts. This is precisely why calibration in physical units matters.

The present preprint should also be interpreted in the context of the CUBS benchmark. Meiburger et al. demonstrated that several classical computerized methods achieved CIMT absolute bias values in the 0.11--0.20 mm range relative to expert A1, with TUMDE and CNR\_IT among the strongest performers~\cite{meiburger2021}. Our DINOv3 system is competitive with that scale, and its main strength lies in providing a modern learned representation, reproducible patient-level evaluation, direct compatibility with qualitative overlays, and an explicit path toward test-time measurement calibration.

\section{Limitations}
This preprint reports a strong segmentation baseline and a calibration-aware CIMT measurement pipeline, but several limitations remain. First, the present manuscript is centered on CUBS v1 only; external generalization to other carotid ultrasound cohorts has not yet been established. Second, although the error scale is competitive with prior automated methods, the correlation across seeds varies, suggesting sensitivity to split composition and image difficulty. Third, our current T3 component is limited to threshold-based test-time calibration rather than a richer self-supervised test-time adaptation of the model itself. Finally, while our literature comparison is meaningful in scale, it is not a strict head-to-head reproduction of all CUBS computerized methods under the exact same experimental pipeline.

\section{Conclusion}
We presented a DINOv3-based framework for carotid intima-media thickness measurement on CUBS v1 that combines segmentation, calibration-aware unit conversion, and test-time threshold calibration. Across three patient-level test splits, the model achieved a mean Dice of 0.7739 and a mean CIMT absolute error of 181.16 $\mu$m. In a held-out validation subset, test-time calibration reduced the image-level CIMT error from 141.0 $\mu$m to 101.1 $\mu$m and reduced the mean bias from 101.3 $\mu$m to near zero. Taken together, these results show that vision foundation models can support interpretable, physically calibrated CIMT measurement, and that test-time calibration targeted at the downstream measurement objective is an effective complement to segmentation-based carotid analysis.

\end{document}